\pdfoutput=1

\documentclass[11pt]{article}

\usepackage[preprint]{coling}

\usepackage{times}
\usepackage{latexsym}

\usepackage[T1]{fontenc}

\usepackage[utf8]{inputenc}

\usepackage{microtype}

\usepackage{inconsolata}

\usepackage{graphicx}

\usepackage{multicol}
\usepackage{multirow}
\usepackage{amsmath}
\usepackage{graphicx}
\usepackage{xcolor,soul}
\usepackage{CJKutf8}
\usepackage{tabularx}
\usepackage{booktabs}
\usepackage{makecell}
\usepackage[misc]{ifsym}
\usepackage{amssymb}
\usepackage{pifont}
\usepackage[export]{adjustbox}

\newcommand{\zhsmall}[1]{\begin{CJK*}{UTF8}{gbsn}\small{#1}\end{CJK*}}

\newcommand{\red}[1]{\textcolor{red}{#1}}
\newcommand{\blue}[1]{\textcolor{blue}{#1}}
\newcommand{\violet}[1]{\textcolor{violet}{#1}}

\definecolor{color1}{RGB}{199,47,47}
\definecolor{color2}{RGB}{242,129,29}
\definecolor{color3}{RGB}{52,157,53}
\definecolor{color4}{RGB}{142,111,173}

\title{Chain-of-Discussion: A Multi-Model Framework for Complex Evidence-Based Question Answering}

\author{Mingxu Tao\textsuperscript{1,2}, \ \ Dongyan Zhao\textsuperscript{1,2,3}, \ \ \textbf{Yansong Feng}\textsuperscript{1\ \Letter} \\
\textsuperscript{1}Wangxuan Institute of Computer Technology, Peking University \\
\textsuperscript{2}Center for Data Science, Peking University \\
\textsuperscript{3}Key Laboratory of Intelligent Press Media Technology, Peking University\\
{\tt \{thomastao,zhaodongyan,fengyansong\}@pku.edu.cn} 
}

\begin{document}
\maketitle
\begin{abstract}
Open-ended question answering 
requires models to find appropriate evidence to form  well-reasoned, comprehensive and helpful answers. 
In practical applications,  models also need to engage in extended discussions on potential scenarios closely relevant to the question. With augmentation of retrieval module, open-source Large Language Models~(LLMs) can produce coherent answers often with different focuses, but are still sub-optimal in terms of reliable evidence selection and in-depth question analysis.  
In this paper, we propose a novel Chain-of-Discussion framework to leverage 
the synergy among multiple open-source LLMs aiming to provide 
\textbf{more correct} and \textbf{more comprehensive} answers for open-ended QA, although they are not strong enough individually. Our experiments show that discussions among multiple LLMs play a vital role in enhancing the quality of answers. 
\end{abstract}

\section{Introduction}






Large Language Models~(LLMs) have demonstrated remarkable language generation capabilities~\cite{brown2020language,touvron2023llama,openai2023gpt4}, propelling advancements in various understanding/generation tasks, including open-domain question answering (QA)~\cite{Song-etal-2024-Evidentiality}. However, for complex open-ended question answering, which plays an important role in human-AI interaction, LLMs may still produce output with hallucination  and often 
deliver inferior performance compared to short-form QA~\cite{huang2023survey}. This task usually requires LLMs to analyze the questions first, retrieve evidence accordingly, then form a long-form answer which is expected to be correct and well-reasoned with details and proper evidence supported. It has a wide range of applications, from legal consultations and medical advice to education support and financial analysis, where users may pose various complex and knowledge-intensive questions. 

\begin{figure*}[!h]
\begin{center}
\centerline{\includegraphics[width=\textwidth]{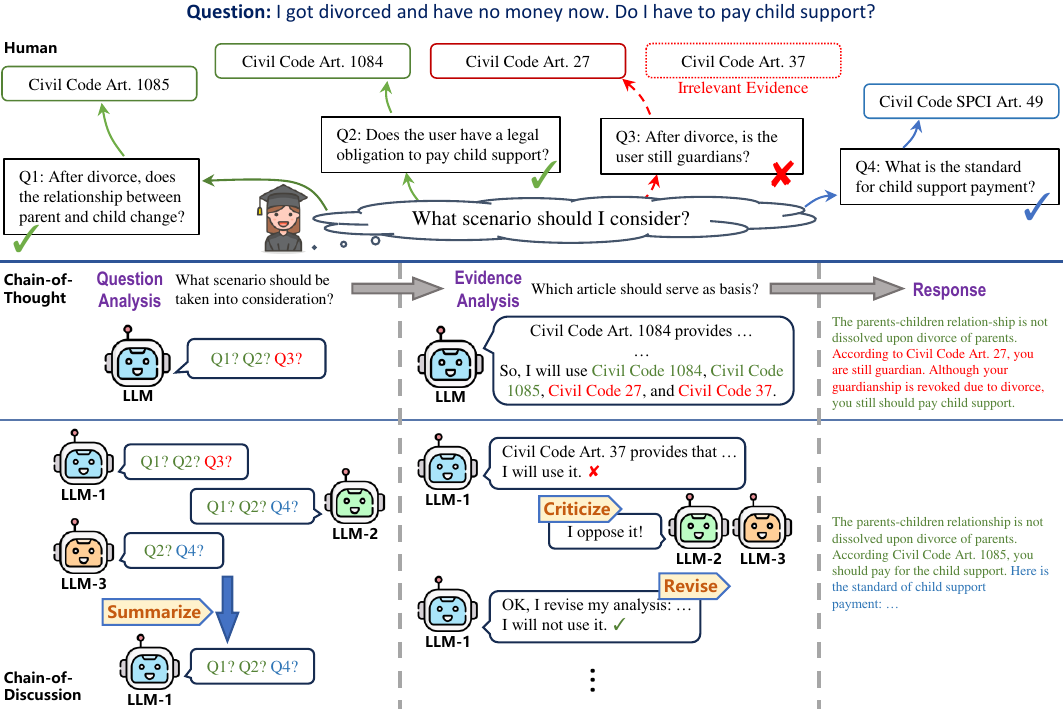}}
\caption{The process of \textbf{Chain-of-Discussion (bottom part)}, compared with chain-of-thought (middle part). The green parts are necessary to answer user's question. Blue parts indicate closely related to the question, useful for detailed/extended discussions. Red parts are irrelevant contents that should be avoided.}  
\label{fig:model} 
\end{center}
\vskip -0.2in
\end{figure*}

Although current LLMs can produce long and coherent texts~\cite{peng2024yarn}, the complex open-ended QA is still an admittedly challenging task, even with augmented retrieval modules. The challenges primarily arise from two aspects. 


Firstly, retrieval models are not entirely prefect, inevitably with noise in the retrieval results. As a legal consultation example shown in Figure~\ref{fig:model}, the model is required to respond to a question regarding the necessity of child support payments. Due to the semantic similarity between obligations for supporting children (financially) and raising/protecting children (physically), the retrieval model may wrongly return law articles pertaining to guardianship qualifications. LLMs usually cannot filter all these noisy evidence, which may propagate and lead to incomplete analysis, wrong reasoning paths, biased opinions and finally problematic or even misleading answers.

Secondly, we expect LLMs output correct responses and consistent explanations, providing useful suggestions about potential scenarios not directly mentioned in questions but indeed helpful for users' current/near-future situations. In Figure~\ref{fig:model}, when responding to a question about the obligation to pay child support for a user facing financial difficulties, the model should also remind her/him of the standards for child support payments and ways to negotiate for a reduction 
given her/his current situation. This is even hard for humans where one should have access to proper evidence, e.g., necessary or closely related law articles, and accordingly provide kind reminders with reasonable explanations. Let alone LLMs without specific training/fine-tuning, which usually focus on the  specific facts literally appearing in the questions.  

In this work, we focus on the complex evidence-based question answering~(CEBQA) task, a typical example of the open-ended QA tasks.
We collect a high quality CEBQA dataset consisting of 200 carefully annotated legal consultation questions in the field of marriage and family affairs.  
To address the challenges, we propose a novel chain-of-thought (CoT) framework,  Chain-of-Discussion~(CoD), encouraging multiple LLMs to summarize, criticize and revise each other's output to reach a well-supported and helpful response. 

Our motivations are two-fold. 
First, different LLMs may have different intrinsic knowledge and reasoning capabilities due to different training data. 
Thus, multiple LLMs can be less possible to make errors concurrently than a single LLM. Recent works~\cite{zhang-etal-2023-sac3} show checking the consistency across multiple LLMs helps reduce output hallucinations. Specifically, we propose a criticize-and-revise framework, which requires multiple LLMs to discuss and reach a consensus for a better response.
For questions that need to involve helpful scenarios or possible extensions, we guess multiple LLMs may provide a diverse set of perspectives to address these possibilities. We thus propose a summarizing step to gather different but helpful perspectives from multiple LLMs, which 
will eventually form comprehensive and detailed responses based on the summarized analyses. 

Different from existing multi-model interaction works~\cite{chan2024chateval,zhang-etal-2023-sac3} using strong closed-source LLMs, e.g., GPT-4~\cite{openai2023gpt4},  
we decide to take a challenge to study how to best exploit the small-scaled open-source LLMs, e.g., around 7B parameters, for a shared objective, while pushing the boundary of research regarding multi-model interaction.

Our main contributions are as follows: (1) We collect 
a high-quality CEBQA dataset consisting of 200 legal consultation questions in Chinese with carefully annotated evidence and answers. 
(2) We propose a novel chain-of-discussion framework, i.e., summarize-criticize-revise, which harnesses the synergy among multiple open-source LLMs to generate more accurate and helpful responses. (3) 
Both GPT-4-based and evidence-centric evaluations demonstrate our framework can help small-scaled LLMs benefit from each other  and improve the overall quality in terms of correctness and comprehensiveness. 

\section{Related Works}
\paragraph{Retrieval-Augmented Generation} \citet{rag} initially propose the paradigm of retrieval-augmented generation~(RAG), which can effectively reduce hallucinations within the texts generated by LLMs. RAG offers a vital solution to mitigate the problem of LLMs lacking domain-specific knowledge, thereby enhancing the credibility of LLMs~\cite{gao2023retrievalaugmented}. In the RAG paradigm, models typically undergo multiple generation steps to achieve the final results. For a user input, models first run a retriever to scan the store of evidence to select several documents as reference. Subsequently, models should determine when and whether to use each evidence document before generating~\cite{izacard2022atlas,shi2023replug,yu2023generate,trivedi-etal-2023-interleaving}. \blue{}

In this work, we face more complex challenges  than RAG. While the model filters out irrelevant evidence, it also needs to retain evidence relevant to potential scenarios. Sometimes, determining which evidence can be used for potential scenarios and which are irrelevant is also challenging for humans.

\paragraph{Chain-of-Thought} Previous works demonstrate that LLMs have a promising capability to decompose a complex question into several intermediate steps~\cite{wei2022chain,kojima2022large}. By segmenting the original question, LLMs can focus on handling each simple sub-question at each step, thus yield more accurate results~\cite{zhou2023leasttomost}. The CoT framework is now widely employed in diverse practical NLP applications~\cite{zelikman2022star,shi2023language,wang2023boosting}. Previous works also employ CoT in the self-correction process of LLMs, which aims to re-generate better outputs. For instance, in Chain-of-Verification, the model generates several queries to verify its original answer, and then revise the answer based on the verification results~\cite{dhuliawala2023chainofverification}. Most of these efforts perform self-checking based on a single model. However, we study a novel CoT framework for multi-model interactive checking and re-generating. 

\section{Preliminaries}

\label{sec:task_def}
\paragraph{Task Definition} 
In CEBQA tasks, given a user's question $q$ and a store of evidence documents $\mathcal{D}$, a model should analyze $q$ first, find necessary evidence $\mathcal{D}_{q}=\{d_1,\ \cdots,\ d_t\}$ from $\mathcal{D}$ accordingly and generate a paragraph $r$ as the final response. 
For instance, in the legal consultation task, users may ask what to do given her/his current situation. The model should find supportive evidence from a store of law articles or previous legal cases, and generate a helpful and detailed response.  


Specifically, 
we expect the generated responses to meet the requirements in terms of correctness and comprehensiveness.
    (1) \textbf{Correctness:} The responses should be based on the evidence that can support to answer the questions, and refrain from employing irrelevant evidence or misinterpreting evidence out of context. 
    (2) \textbf{Comprehensiveness:} The responses should engage in discussions about potential scenarios that would be relevant/helpful to users, even not explicitly mentioned in questions. 

We note that it is hard to guarantee all the retrieved evidence pieces can be perfectly used to answer the question. Therefore, similar to RAG, models should filter out irrelevant evidence. However, it is more challenging for models to carefully retain the evidence that can be used for discussions about potential scenarios, even though the evidence may not directly support answering the question.




\paragraph{Baseline Framework: CoT}
Previous works have revealed that the CoT  prompt can enhance the ability of LLMs to handle complex reasoning tasks~\cite{wei2022chain,kojima2022large}. Inspired by these works, we employ a 
multi-step prompt to stimulate LLMs to generate more correct while comprehensive answers. 

We initially prompt LLMs to analyze the question $q$, including identifying the possible role of users, understanding explicit and implicit demands of users, and determining what types of evidence is needed to answer the question. The generated \underline{a}nalysis of question can be denoted as $a^{\text{que}}_{q}$.  

The next step is to judge whether each evidence document can serve as a potential basis for responding to the question $q$. Here, we employ a prompt to feed the LLM with question $q$, analysis $a^{\text{que}}_{q}$ of the question, and a specific evidence document $d_i$. 
The LLM then need to analyze whether $a^{\text{evi}}_{d_i}$ can be used to address the issues raised in $q$
and whether evidence $d_i$ can probably be used to respond or not.

The LLM with parameters $\theta$ should finally respond to the question $q$ according to question analysis $a^{\text{que}}_{q}$ and evidence analysis $\left\{a^{\text{evi}}_{d_i}\right\}_i$, based on the evidence document set $\mathcal{D}_q$:
$$r=f\left(q, \mathcal{D}_{q}, a^{\text{que}}_{q}, \left\{a^{\text{evi}}_{d_1}, \cdots, a^{\text{evi}}_{d_t}\right\}|\theta\right).$$

As observed in our pilot study, one small-scaled LLM could generate fluent answers, but often with incomplete analysis or  wrong reasoning paths. 

\section{CoD: Summarize, Criticize, and Revise}

Our Chain-of-Discussion framework leverages interactive discussions among multiple LLMs, thereby addressing potential shortcomings in individual's intrinsic knowledge. 

Similar to the baseline, we employ a two-stage analyzing pipeline that 
instructs LLMs to analyze the question and evidence separately. 
To address the correctness and comprehensiveness  of generated answers, during question analysis, we encourage models to read and summarize others' analyses so as to consider more scenarios closely relevant to the question, in the purpose of augmenting the comprehensiveness. 
During the stage of evidence analysis, we require all other LLMs to \textbf{criticize} the evidence analysis of each LLM. Subsequently, the model will read others' critique and determine whether to \textbf{revise} its own analysis or not.
The model finally generate a correct and more helpful response based on the summarized question analysis and revised evidence analysis.

\subsection{Stage 1: Question Analysis}

Formally, suppose there are $n$ accessible LLMs, denoted as $M_1,\ \cdots,\ M_n$. For a given question $q$ and the retrieved evidence $\mathcal{D}_q$, we aim to employ the target LLM $M_k$ to generate a response, with the assistance of the remaining LLMs. 

We first instruct the LLMs to analyze the question, including facts mentions in $q$, primary needs of the user, and potential scenarios 
associated with the question. 
We observe that LLMs may perform poorly in analyzing potential scenarios when solely 
relying on their intrinsic knowledge, especially those models that have not been pre-trained or supervised fine-tuned on domain-specific data. Thus, we argue that the evidence documents $\mathcal{D}_q$ can serve as vital cues about the potential scenarios not mentioned in $q$.

Different LLMs can have varying preferences in analyzing the potential scenarios. Therefore, we believe that by 
integrating the outputs of multiple LLMs, we can 
take more helpful scenarios into account, thus 
improve the \textbf{comprehensiveness} of question analysis. We prompt each LLM $M_i$ to analyze the question $q$, with retrieved evidence $\mathcal{D}_q$ as a reference: $a_{q,\,M_i}^{\text{que}}=f_{\text{que}}\left(q,\ \mathcal{D}_q \vert \theta_{M_i}\right).$

We then employ the target LLM $M_k$ to \textbf{summarize} the question analyses of all models, according to following instructions:
    
    $\bullet$ \textbf{Consistency:} If the majority of LLMs provide similar analyses regarding a fact in the question or a potential scenario, then it is likely to be correct. You can include it in the summary.
    
    $\bullet$ \textbf{Comprehensiveness:} If a minority of LLMs hold a particular viewpoint in their analyses with reasons, it does not imply its unreliability. You should scrutinize this content, assessing its logical coherence and relevance to the question.

The summarized question analysis can be $a^{\text{que}}_q=f_{\text{sum}}\left(q,\ a^{\text{que}}_{q,\,M_1},\ \cdots,\ a^{\text{que}}_{q,\,M_n} \vert \theta_{M_k}\right)$.

\subsection{Stage 2: Evidence Analysis}
\label{sec:evi_analysis}

Incorporating many irrelevant evidence documents as  input would inevitably introduce noise, which could deteriorate the model performance. Thus, we should discern which evidence document should be used to address the question. For an evidence document $d_j\in \mathcal{D}_q$, we prompt the target model $M_k$ to analyze it based on the question and question analysis : $\hat{a}_{d_j}^{\text{evi}}=f_{\text{evi}}(d_j,\ q,\ a^{\text{que}}_{q}\vert \theta_{M_k})$.

However, a single LLM might generate hallucinated outputs~\cite{li-etal-2023-halueval,huang2023survey}, and incorrectly assess the relevance between  
evidence documents and the given question. Inspired by previous work~\cite{zhang-etal-2023-sac3}, we propose a multi-party discussion framework to improve the quality of evidence analysis. 

First, we instruct each LLM, excluding $M_k$, to \textbf{criticize} the evidence analysis $\hat{a}_{d_j}^{\text{evi}}$. 
Each critic model $M_i$ should explicitly output whether it holds opinions contrary to $\hat{a}_{d_j}^{\text{evi}}$, which are denoted as $c^{d_j}_i$. In this work, we employ a revising threshold $\delta$. 
If the proportion of opposite opinions in the critiques exceeds $\delta$, the target model needs to \textbf{revise} its evidence analysis: $a^{\text{rev}}_{d_j}=f_{\text{rev}}\left(q,\ d_j,\ a^{\text{que}}_q,\ \hat{a}_{d_j}^{\text{evi}}|\{c_i^{d_j}\}_{i},\ \theta_{M_k}\right).$ %

We assume that the critique requiring to revise can be reliable only when a majority of critic models achieve a consensus. Otherwise, we retain the original evidence analysis. Formally, we collect the evidence analysis as following:
$$
a_{d_j}^{\text{evi}}=
\begin{cases}
\hat{a}_{d_j}^{\text{evi}}, & \text{if $\frac{\left\vert\{c_i|c_i=\text{opposite}\} \right\vert } {\left\vert\{c_i\}\right\vert } \leq \delta$;}\\
a^{\text{rev}}_{d_j}, & \text{otherwise}.
\end{cases}
$$

\subsection{Response Generation}

For a fair comparison, we employ prompts similar to those of the baseline framework to generate responses. 
We denote the response as $r=f_{\text{ans}}\left(q, \mathcal{D}_{q}, a^{\text{que}}_{q}, \left\{a^{\text{evi}}_{d_1}, \cdots, a^{\text{evi}}_{d_t}\right\}|\theta_{M_k}\right).$

\section{Experiments}
As there is no existing dataset for CEBQA tasks, here we delve into the legal consultation task 
and collect the first CEBQA dataset.
In civil law systems like China, all legal activities, including legal consultation,  should be based on \textit{law articles} (or \textit{judicial interpretations}), which can be naturally considered as the evidence store in our framework.

\subsection{Data Collection}
\label{sec:data_collection}
To reflect the diversity in practical scenarios, we focus on the fields of \textit{marriage, family affairs, and inheritance}, covering various legal disputes, e.g., \textit{divorce, custody, contracts, property}, etc. 
From a pool of 609 questions collected from real users with consultants' responses through Web Search Engines, we employ an legal expert to select 200 questions\footnote{To access the  legal consultation data: \url{https://github.com/kobayashikanna01/Chain-of-Discussion}} to ensure their semantic distinctiveness and coverage. 
See more details in Ethics Statement.

\paragraph{Evidence Annotation} 
We construct the evidence store based on all  1,749 articles of the \textit{Civil Code}, \textit{Civil Procedure Law} and their judicial interpretations in China. For each question, we consider three types of articles: \textit{necessary}, \textit{optional}, and \textit{not required}. \textit{Necessary} articles are highly relevant to the question, while \textit{optional} ones can be basis for discussions of potential scenarios (more details in Appendix~\ref{app:evi_anno}). We ensure there are 5 articles retrieved for each question, and on average, each contains 1.52 \textit{necessary}, 1.23 \textit{optional}, and 2.25 \textit{not required} articles. 
It means about 45\% of the retrieved articles are not required at all. 


\paragraph{Data Quality} 
We employ 6 annotators with background in civil law to manually check the questions, answers and articles. 
Annotators are instructed to correct all typos but retain the informal expressions in questions. 
Note that there are many omissions or slight word-order inversions in the questions, posing a challenge to models' reasoning capabilities.

Annotators also examine the correctness and logical coherence of the responses. 
For problematic ones,  annotators are encouraged to discuss and reach a consensus 
for modifications, otherwise, leave them as they are.
Averagely, it takes about 20 hours per annotator to examine 100 instances.

\subsection{Experimental Setup}

As our main focus is to investigate whether small open-source LLMs can collaborate through summarize-criticize-revise, 
we study four open-source fine-tuned LLMs, Baichuan2-7B~\cite{baichuan2023baichuan2}, Deepseek-7B~\cite{deepseek-llm}, Qwen-7B~\cite{bai2023qwen}, and Xverse-7B\footnote{\url{https://huggingface.co/xverse/XVERSE-7B-Chat}}, which are four of the best 7B-parameter LLMs performing on CMMLU~\cite{li2023cmmlu}. When we use a specific LLM as the target model, the other three LLMs are expected to generate diverse question analyses and criticize the evidence analysis of target model. 

To examine if close-source LLMs can still benefit from our CoD, we also test with \textit{gpt-3.5-turbo-1106}, \textit{gemini-1.0-pro-latest}, and \textit{claude-3-haiku-20240307} similarly to the open-source group. 

We note that the two stages in Chain-of-Discussion are independent of each other. Therefore, we can investigate 
how they contribute to the ultimate performance
by the following settings: 
    
    \noindent\textbf{Single-model baselines~(BS):} One LLM takes a query and several retrieved articles as input and performs question analysis, article analysis, and response generation in a vanilla CoT manner.




    
     \noindent\textbf{Only Stage 1~(S1):} All LLMs produce question analysis. The target LLM summarizes these analyses, and proceeds to the rest by itself. 
    
     \noindent\textbf{Only Stage 2~(S2):} Three other LLMs criticize the article analysis generated by the target LLM. The question analysis and the final response are generated by the target LLM on its own.
    
     \noindent\textbf{Chain-of-Discussion~(S1S2):} All LLMs involve into both question analysis and article analysis. Eventually, the target LLM produces the response by itself.
%

 
  We employ each LLM as the target model, replicating the experimental settings. 
  We report the performance for each LLM as the target role. 
 More details and prompt templates are in Appendix~\ref{app_gen_detail} and ~\ref{app:prodis}.

\paragraph{Evaluation Metrics} 
Different from conventional QA tasks, 
the responses in CEBQA tasks can consist of several hundred or even thousands of words, which are also knowledge intensive and complex in structures, containing facts and causal relations to be verified. Therefore, it is impossible to employ the popular metrics such as F1 or exact match~\cite{joshi-etal-2017-triviaqa,rajpurkar-etal-2018-know}. 

Following previous works~\cite{liu-etal-2023-g,chan2024chateval}, we employ GPT-4 to evaluate the quality of generated responses, with the expert-written responses, necessary and optional articles as reference. We prompt \texttt{gpt4-turbo-0125} to score a response in an integer between 1 and 10 based on correctness and comprehensiveness. 
If there is no clear reason to indicate a response  is significantly better or worse than human-written ones, a score of around 7 should be given (Scoring prompt is in Appendix~\ref{app:gpt4_scoring}).

\subsection{Main Results}

\begin{table}[t]
\small
\centering
\begin{tabular}{l|lrr}
\toprule
\textbf{Target LLM} & \textbf{Setting} & \textbf{Avg. Score} & \textbf{$\Delta$Score}\\
\midrule
\multicolumn{4}{c}{Open-source LLMs Group}\\
\midrule
\multirow{4}*{Baichuan2-7B} & BS & 5.750 & --\\
& S1 & 6.030 & +0.280\\
& S2 & 5.935 & +0.185\\
& S1S2 & \textbf{6.090} & +0.340\\
\midrule
\multirow{4}*{Deepseek-7B} & BS & 6.465 & --\\
 & S1 & 6.505 & +0.040 \\
 & S2 & 6.480 & +0.015 \\
 & S1S2 & \textbf{6.580} & +0.115\\
\midrule
\multirow{4}*{Qwen-7B} & BS & 5.835 & --\\
 & S1 & 5.890 & +0.055 \\
 & S2 & 5.815 & -\,0.020\\
 & S1S2 & \textbf{5.955} & +0.120\\
\midrule
\multirow{4}*{Xverse-7B} & BS & 6.015 & --\\
 & S1 & 5.995 & -\,0.020\\
 & S2 & 6.030 & +0.015\\
 & S1S2 & \textbf{6.125} & +0.110\\
 \midrule
\multicolumn{4}{c}{Close-source LLMs Group}\\
\midrule
\multirow{2}*{GPT-3.5-turbo} & BS & 6.895 & --\\
& S1S2 & \textbf{6.955} & +0.060\\
\midrule
\multirow{2}*{Gemini-1.0-pro} & BS & 6.940 & --\\
 & S1S2 & \textbf{6.975} & +0.035\\
\midrule
\multirow{2}*{Claude-3-haiku} & BS & 7.280 & --\\
 & S1S2 & \textbf{7.300} & +0.020\\
\bottomrule
\end{tabular}
\caption{The average scores of each target LLM and each setting evaluated by GPT-4. The upper is evaluated within the open-source LLMs group, while the bottom is within the closed-source LLMs group. }
\label{tab:gpt4_scores}
\end{table}

Table~\ref{tab:gpt4_scores} shows the evaluation results produced by GPT-4. Comparing the baseline CoT setting~(BS) and Chain-of-Discussion~(S1S2) in the open-source group, we can find \textbf{each LLM can obtain improvements from discussions with other LLMs}, with Baichuan2-7B increased by +0.340, Deepseek-7B by +0.115, Qwen-7B by +0.120, Xverse-7B by + 0.110. 
 We think other LLMs bring more related aspects according to their own strengths into discussions while the baseline COT setting has to rely on one LLM only. 
We also find employing multi-model discussion on both stages can bring more improvement than using it on one stage only. 


Although CoD can enhance all LLMs,  the CoD-augmented Baichuan2-7B, Qwen-7B, or Xverse-7B still can not outperform Deepseek-7B under its baseline setting, with around 0.5 left behind.
This may indicate that the quality of responses primarily relies on the inherent ability of LLMs to analyze context and then to generate. 

As shown in the bottom of Table~\ref{tab:gpt4_scores}, 
even for powerful closed-source LLMs like Claude-3, our CoD framework can still brings improvement consistently, but the improvements are admittedly small compared to the open-source group.

Through a manual check, we also find that these powerful closed-source LLMs may also produce erroneous analyses of questions, resulting in incorrect responses. We believe that regardless the model scales, our CoD framework can leverage the varied knowledge and capabilities of multiple models to achieve better performance.

We provide more results for other baseline methods in Appendix~\ref{app:other_baseline}.

\section{Discussions}
\label{sec:discussion}
\vspace{-0.3em}
\subsection{Evidence-Centric Evaluation}

Besides overall evaluation by GPT-4,  we wonder if our Chain-of-Discussion framework can enhance the comprehensiveness and correctness of the model output.
%
When discussing the details of questions or potential scenarios, 
LLMs should refer to \textit{necessary} or \textit{optional} evidence. Hence, we can assess the correctness and comprehensiveness of responses by the accuracy of evidence documents. 

We accordingly design two metrics of accuracy, \texttt{N-Acc} and \texttt{O-Acc}, to assess the correctness and comprehensiveness, respectively. We utilize the \textit{not required} articles as negative samples. For \texttt{N-Acc}, we employ the \textit{necessary} articles as positive samples, while the \textit{optional} articles for \texttt{O-Acc}. We employ heuristic methods to examine if a response has used an article (details in Appendix~\ref{app:art_rule}).

\begin{table}[t]
\small
\centering
\begin{tabular}{l|lcc}
\toprule
\textbf{Target LLM} & \textbf{Setting} & \textbf{\texttt{N-Acc}\%} & \textbf{\texttt{O-Acc}\%}\\
\midrule
\multirow{4}*{Baichuan2-7B} & BS & 58.26 & 50.14 \\
& S1 & 60.03 & \underline{50.67} \\
& S2 & \underline{61.86} & 50.25 \\
& S1S2 & \textbf{63.17} & \textbf{52.38} \\
\midrule
\multirow{4}*{Deepseek-7B} & BS & 75.93 & 59.27 \\
 & S1 & \underline{76.36} & \underline{59.70} \\
 & S2 & 76.12 & 59.23 \\
 & S1S2 & \textbf{76.79} & \textbf{59.80} \\
\midrule
\multirow{4}*{Qwen-7B} & BS & 69.87 & 60.98\\
 & S1 & 70.31 & 61.63 \\
 & S2 & \underline{70.64} & \underline{63.65} \\
 & S1S2 & \textbf{71.29} & \textbf{64.20} \\
\midrule
\multirow{4}*{Xverse-7B} & BS & 74.00 & 63.95 \\
 & S1 & 74.24 & \underline{64.72} \\
 & S2 & \underline{75.67} & 64.44 \\
 & S1S2 & \textbf{76.16} & \textbf{65.35} \\
\bottomrule
\end{tabular}
\caption{The Macro average \texttt{N-Acc} and \texttt{O-Acc} results of each target LLM and each setting. The highest scores are made \textbf{bold}, while the second \underline{underlined}.}
\label{tab:evi_centric}
\end{table}

We compute the Macro average \texttt{N-Acc} and \texttt{O-Acc} across all examples, shown in Table~\ref{tab:evi_centric}.  
We can see these LLMs  exhibit differently on their own, and when performing a simple majority vote among all LLMs, we could get 77.62\% for \texttt{N-Acc}, confirming our hypothesis that \textbf{it is possible to harness the synergy among multiple open-source LLMs for a better performance}.
Compared to the baselines~(BS), our CoD~(S1S2) can achieve around a 2\% improvement on both \texttt{N-Acc} and \texttt{O-Acc} for Baichuan2-7B, Qwen-7B, and Xverse-7B. Even for Deepseek-7B, the best in GPT-4 based evaluation, our  framework still brings improvements of 0.86\% and 0.53\% to \texttt{N-Acc} and \texttt{O-Acc}, respectively. %


The results indicate that introducing multi-model discussions during both question analysis and evidence analysis \textbf{helps target LLMs better refer to correct evidence}. This also explains why CoD 
can improve the quality of model responses.

Comparing the results under BS, S1, and S2, we find that involving multiple LLMs in a single stage can actually enhance both correctness and comprehensiveness. Specifically, employing multi-model discussions in question analysis contributes more to comprehensiveness, while introducing other models in evidence analysis (BS vs. S2) helps filtering irrelevant evidence thus brings more improvement in correctness. 


\subsection{Human Evaluation}

We further examine the quality of generated responses through a win-rate analysis by human experts. We randomly sample 50 examples, and ask two human experts to examine whether the responses of CoD-enhanced models are better than those of baselines. We use the responses of Baichuan2-7B and Xverse-7B, which are models obtaining the most and the least improvement from CoD, respectively.



\begin{figure}[t]
\begin{center}
\centerline{\includegraphics[width=0.9\columnwidth]{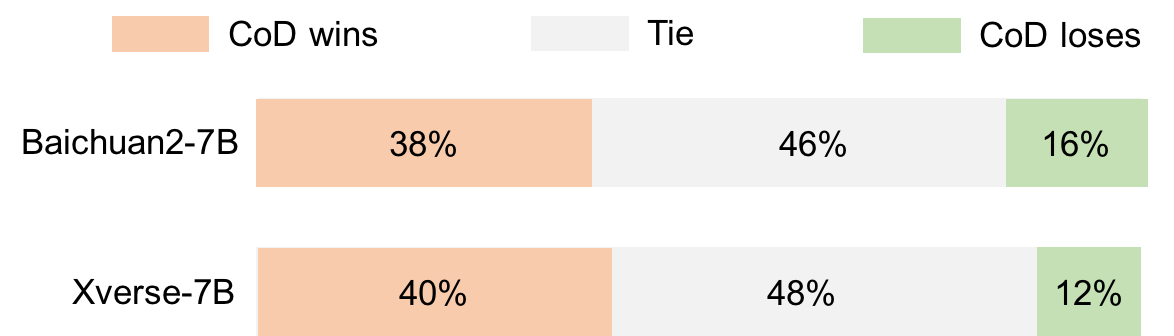}}
\caption{Human preference evaluation, comparing the CoD settings of Baichuan2-7B and Xverse-7B  to their corresponding baseline settings across 50 randomly sampled examples.}  
\label{fig:human} 
\end{center}
\vskip -0.2in
\end{figure}

As shown in Figure~\ref{fig:human}, in 38\% of the examples, CoD can bring improvements to the responses of Baichuan2-7B, while 40\% responses generated by Xverse+CoD surpass Xverse's baseline setting. For Baichuan2-7B, only 16\% of responses deteriorated after introducing the CoD framework, while for Xverse, only 12\% worsened.

\noindent\textbf{Qualitative Analysis.} We also manually examine 30 responses of both models. The main error types are \textit{misunderstanding key legal terms} (60\%), \textit{discussing not-required articles} (15\%), and \textit{inconsistent explanations} (10\%). The latter is often caused by term misunderstanding.
This shows that there may not be sufficient legal knowledge in 7B models to  distinguish very similar terms. Comparing our CoD and baseline setting, we can see CoD can correct improper articles for 6 out 30 cases, leading to better results, through the cooperation mechanism. 

\noindent\textbf{Correlation between GPT-4 Evaluation and Human} Although GPT-4 has been deemed as a reliable evaluator in various LLM evaluations~\cite{liu-etal-2023-g}, we still want to examine how well GPT-4 preferences correlate to human experts in our challenging task. 
For each example from the randomly sampled 50 cases:
(1) if GPT-4 assigns a higher score for the response generated by the CoD setting, we denote this evaluation result as +1;
(2) if GPT-4 assigns a higher score for the response by the baseline setting, we denote this evaluation result as -1;
Otherwise, we denote it as 0.
We use the same principle to annotate the evaluation results of human expert.

The Pearson correlation coefficient between GPT-4 scores and human expert evaluation is 0.5863 with  $p=7.7\times 10^{-6}$. This indicates a relatively high correlation between the two evaluation methods, where we could consider GPT-4 evaluation as a reliable manner for our work.

\subsection{Limitations of Open-Source LLMs}
\label{sec:llm_limitations}
%
%

As shown in Table~\ref{tab:gpt4_scores}, using multi-model interaction at only one stage fails to enhance Xverse-7B (S1) or Qwen-7B (S2). 
We guess this may be due to the small size of the model scales, 7B here. 


After a manual check, we find the main difficulties for those small scaled open-source LLMs: 
1) those 7B LLMs may not have sufficient legal/background knowledge and reasoning capabilities, thus often fail to discern errors in the analysis of others and integrate them into the summary; 
2) the small scaled LLMs do not always follow instructions, and often refuse to modify the errors in evidence analysis. 



\begin{table}[t]
\centering
\begin{small}
\begin{tabular}{p{0.97\columnwidth}}
\toprule
\textbf{Question:} Can a daughter/son still be eligible to inherit if she/he has not supported her/his parents?\\
\midrule
\textit{Original question analysis:} \\
\textbf{Baichuan2:} This question involves the relationship between inheritance rights and filial duties. The user may be concerned about whether a person who has not fulfilled filial duties still qualifies for inheritance. \violet{This may entail rules regarding the recognition of inheritance rights, the forfeiture and reinstatement of inheritance rights, and the role of filial duties in the determination of inheritance rights.}\\
\midrule

\sethlcolor{yellow}\textit{Question analyses of other LLMs:}\\ \textbf{Qwen:} ... \ \ \ \ \ \ \ \ \ \ \  \textbf{Xverse:} ...\\ \textbf{Deepseek:} ... \violet{The user may want to clarify whether not supporting parents will affect her/his eligibility to inherit parents' estate.} Additionally, user may also inquire about \hl{her/his entitlement to demand child support payments from her/his parents, if the parents refuse to raise children.} ... \\
\midrule
\textit{Summarized question analysis:}\\ 
\textbf{Baichuan2:} This question involves the relationship between inheritance rights and filial duties. The user may wonder if her/his child retains eligibility for inheritance despite failing to fulfill filial duties. \violet{This may entail rules regarding the recognition of inheritance rights, the forfeiture and reinstatement of inheritance rights, and the role of filial duties in the determination of inheritance rights.} \hl{The user may also want to know the regulations regarding whether parent who does not raise children is required to pay child support.}\\
\bottomrule
\end{tabular}
\caption{\label{tab:bad_summary_case}
An example of hallucinated summary of question analysis. \violet{Violet texts} are correct analysis, while the \hl{texts with yellow background} are hallucinated parts.
}
\end{small}
\end{table}

\noindent\textbf{Hallucination Propagation in Summary.}
Table~\ref{tab:bad_summary_case} illustrates an example where Baichuan-7B incorporates inaccurate analysis from DeepSeek-7B into the summary.
We find that both LLMs can correctly identify the primary intention 
from the question, i.e., determining whether providing support for parents is a prerequisite for inheriting. Unfortunately, Deepseek-7B mistakenly engages in analysis regarding whether parents pay support for children, while Baichuan2-7B fails to identify this as irrelevant to the question.

Although both raising children and supporting parents are financial support and life care between two generations in a family, it is a commonsense, for people with legal background, that these are two different concepts and legally, there is no causal relationship between them. We think both LLMs actually rely on shallow semantic similarity to judge if a question analysis is relevant to user's question.

\noindent\textbf{Capturing Nuance between Related Terms. }
We guess that open-source LLMs with around 7B parameters may not have sufficient knowledge and enough reasoning capabilities to distinguish related but different terms. Table~\ref{tab:human_case} shows an example 
where Qwen-7B fails to recognize that \zhsmall{复婚} (reconcile and remarry) 
and \zhsmall{再婚} (remarry with other person) 
represent different marriage relationships, comply with different articles or provisions, and should not be used interchangeably.   
However, in our CoD setting,  the discussions with other models can, to some extent, have the chance to reduce the hallucination caused by the target model's poor reasoning capabilities, thus help to reduce logical errors in Qwen-7B's responses.  

\begin{table}[t]
\centering
\begin{small}
\begin{tabular}{p{0.97\columnwidth}}
\toprule
\textbf{Question:} What is the difference between \textit{resumption of marital relationship}~(\zhsmall{复婚}) and \textit{remarriage with other person}~(\zhsmall{再婚})?\\
\textbf{Article:} \textit{Article 1046} A man and a woman shall enter into marriage freely and voluntarily. ...\\
\textit{Article 1083} Where, after divorce, both the man and the woman voluntarily intend to resume their marital relationship, they shall file for re-registration...\\
\midrule
\textbf{Qwen-7B:} ... According to Article 1083, both parities should be voluntary for \textit{resumption of marital relationship}, while \hl{there is no such limitation for \textit{remarriage with other person}.} ...\\
\textbf{Qwen-7B+CoD:} ... According to Article 1046, \violet{whether it is \textit{Fuhun} or \textit{Zaihun}, both parities need to do so voluntarily.}\\
\bottomrule
\end{tabular}
\caption{\label{tab:human_case}
A case of Qwen-7B obtaining improvement from CoD. \violet{Violet texts} are correct analysis, while the \hl{texts with yellow background} are hallucinated parts. 
}
\end{small}
\end{table}


\noindent\textbf{Reasoning with Knowledge. }
Similarly, we find the lack of good reasoning capabilities and knowledge may also limit the performance of LLMs in revising evidence analysis. LLMs tend to focus on the literal differences between questions and law articles.  
Users often omit commonsense information in their questions, whereas law articles exhibit a rigorous and detailed structure. The stylistic differences between questions and evidence may lead to the misjudgments of LLMs (more examples and analysis in Appendix \ref{app:analysis_of_commonsense}). 

\subsubsection{Model Preference of Revising}

We find different LLMs have varying preferences regarding whether to modify the evidence analysis. 

To quantitatively study this issue, we initially employ Baichuan2-7B to generate the original evidence analyses. Deepseek-7B and Xverse-7B then play the role of critics to determine which analysis should be revised. Then, we use Baichuan2-7B and Qwen-7B to revise these analyses, respectively.  
We find that Baichuan2-7B successfully revises 96.5\% of the analyses, while Qwen-7B can only revise 56.1\% of them.

To mitigate the influence of target model selection, 
we also use Qwen-7B to provide origianl evidence analyses, still with Deepseek-7B and Xverse-7B as the critics. Similarly, Baichuan2-7B can revise 92.5\% of the analyses, but Qwen-7B only revise 67.2\% of them.

We argue that an LLM's preference for refusing to revise may lead to a failure to obtain better evidence analysis based on the critiques. Consequently, it might limit our Chain-of-Discussion framework to bringing more improvement as expected. The preference of LLMs can be affected by supervised fine-tuning and reward modeling~\cite{ouyang2022training,rafailov2023direct}. We hope to study the effect of supervised training on Chain-of-Discussion 
in future.

\section{Conclusions}

In this work, we proposed a novel reasoning framework, Chain-of-Discussion, for complex evidence-based question answering tasks. The CoD framework involves multiple LLMs in discussions to achieve more correct and comprehensive responses with less hallucination and more supportive evidence. Experiments on a challenging legal consultation dataset show CoD can effectively improve the performance of open-source LLMs by encouraging them to discuss and criticize. 

\section*{Limitations}


Our proposed framework is designed to generate correct and comprehensive answers to respond complex questions. When used for providing legal advisory services, this technique can produce helpful responses to help people with needs, but it still cannot guarantee all responses are completely correct. Hence, this techniques should be used with cautions for further applications.

Our dataset is designed and annotated to reflect the nature of CEBQA tasks, which requires models to generate detailed analysis to each closely relevant scenarios of the user's question. However, our annotated results may be inevitably not perfect from the professional perspectives of experts in civil law. Thus it should be used with caution and for research purpose only.

We also note that the proposed framework involves multiple LLMs to generate for several rounds. Straightly using commercial APIs may lead to more promising generated results and cost less time. However, our aim is to validate how to better and more efficiently exploit the synergy among small LLMs, without relying on larger LLMs. We pioneer to expand the border of investigation about multi-model interaction to the small open-source LLMs.

\section*{Ethics Statement}

In this work, we collect the legal consultation question-answer pairs through Web Search Engine. 
All these data are publicly accessible on the internet. 
Additionally, these websites all allow crawlers of search engines to automatically access their web pages.

To protect the privacy of the consultants in real world, we manually check all the instances, and remove those which may indicate the name, address, phone number of a specific users.
Note that information such as gender, age, and family composition serves as crucial background in judicial practice. 
Since this information does not identify a specific individual and does not result in privacy leaks, we do not discard data where the users mention their gender, age, or family composition.

\section*{Acknowledgements}
This work is supported in part by NSFC (62161160339) and Beijing Science and Technology Program (Z231100007423011).
We thank the anonymous reviewers for their helpful discussions and suggestions. For any correspondence, please contact Yansong Feng.

\bibliography{anthology,custom}

\appendix

\section{Details of Generation}
\label{app_gen_detail}

During generation, we set the temperature to 0.3, the repetition penalty to 1.05, and the top-p to 0.8. At Stage 2, the revising threshold $\delta$ is set to 0.66. To enhance the quality of model outputs, we employ in-context learning~\cite{brown2020language}. For each step, due to the limit of context length, we construct 2 human-annotated examples for in-context learning. We ask LLMs to regenerate for fragmentary outputs. 

\section{Details of Evidence Annotation}
\label{app:evi_anno}

Following previous work~\cite{huang2023lawyer}, we train a classification model to retrieve relevant articles. We fine-tune RoBERTa-large~\cite{roberta} with 80K examples. 
Each example consists of one question and 1\textasciitilde 5 articles. For each consultation question, we keep the top 10 articles with the highest probability scores predicted by the classifier. 

To avoid the model failing to retrieve articles that should serve as the basis, we employ 6 annotators with legal backgrounds to supplement missing articles. They are then asked to assess whether each article should/can serve as the basis to respond the given question based on the following principles:
\begin{itemize}
    \item \textbf{Necessary:} The article is highly relevant to the question. Without this article, it is impossible to answer the question.
    \item \textbf{Optional:} The article pertains to potential scenarios that may be relevant to the question. This article can be used for extended discussions about the scenarios that user may concern or face to in near future.
    \item \textbf{Not required:} The article is entirely unnecessary to address the question.
\end{itemize}
We assign relevant scores of 2, 1, and 0 to the three categories of articles, respectively. If the average score of an article exceeds 1.66, it will be regarded as a \textit{necessary} one. And the articles with average scores less than 0.67 can be regarded as \textit{not required}, while the remaining ones are \textit{optional}. 

To imitate retrieval-augmented generation, 
we provide five articles for each question, supposing them as retrieval results. We first keep all the \textit{necessary} and \textit{optional} articles. We then select \textit{not required} articles in descending order of the probability scores predicted by the classifier.

\section{Extended Evaluation}
\label{app:other_baseline}

\subsection{Results of Other Baselines}

In this work, we carefully design a two-stage framework for the legal consultation task, including question analysis and evidence analysis. To give a fair comparison, the chain-of-thought baseline also follows the two-stage settings for generation. 

In our early attempt, we also implement a \textit{real-vanilla} chain-of-thought baseline~\cite{wei2022chain}, which only asks the LLM to directly think whether an article can be used as the evidence and then output the response. 

As an ablation study, we also implement a self-criticism baseline~\cite{tan-etal-2023-self}. Compared to the mutual-criticism method in Section~\ref{sec:evi_analysis}, we only ask the target LLM to review whether to revise its evidence analysis, without the help of other LLMs.

Table~\ref{tab:more_results} shows the results of these two baselines. 
We find the \textit{real-vanilla} CoT baseline demonstrates much worse performance for all four LLMs. Without the step of question analysis, the 7B LLMs fail to select correct articles, and such failure results in erroneous responses.
We can also find a 7B LLM cannot perform well to criticize itself. As we discuss in Section~\ref{sec:llm_limitations}, the 7B LLMs sometimes fail to provide useful insights about revising its own inaccurate analyses. It is also one of our motivations to introduce the collaboration among multiple models.

\begin{table}[t]
\small
\centering
\setlength\tabcolsep{3pt}
\begin{tabular}{lcccc}
\toprule
\textbf{Method} & \textbf{Baichuan2} & \textbf{Deepseek} & \textbf{Qwen} & \textbf{Xverse} \\
\midrule
Vanilla CoT & 5.490 & 6.360 & 5.545 & 5.800\\
Self-Criticism	& 5.640 & 6.250 & 5.790 & 6.005\\
\midrule
Two-Stage CoT & 5.750 & 6.465 & 5.835 & 6.015\\
Our CoD & \textbf{6.090} & \textbf{6.580} & \textbf{5.955} & \textbf{6.125}\\
\bottomrule
\end{tabular}
\caption{The average scores of each method with distinct target LLMs (evaluated by GPT-4).}
\label{tab:more_results}
\end{table}

\subsection{Results under More Evidences}

In the main experiments, we provide five articles for each query, including 2.75 required articles on average. 
We examine whether Chain-of-Discussion can be effective when provided more candidate articles. Thus, we add extra articles, changing the number of given evidences to 7 or 10. 

Table~\ref{tab:num_of_evi} shows the results when we provide 5, 7 or 10 articles as evidence, respectively. As the number of articles increases, the qualities of the responses generated by both baseline and Chain-of-Discussion become a bit worse. It may be attributed to the noise brought by more irrelevant articles, which results in more incorrect contents in model’s responses.

We also find \textbf{when the number of given articles increases, Chain-of-Discussion can bring more improvement} to both Baichuan2-7B and Deepseek-7B, which are the worst-performing and the best-performing models in this work. We believe that Chain-of-Discussion can better filter irrelevant evidence from the given articles.

\begin{table}[t]
\small
\centering
\setlength\tabcolsep{3pt}
\begin{tabular}{lccc}
\toprule
\textbf{Num. of Articles} & \textbf{5} & \textbf{7} & \textbf{10} \\
\midrule
Baichuan2+CoT & 5.750 & 5.680 & 5.575\\
Baichuan2+CoD & 6.090 & 6.075 & 6.030\\
$\Delta(\text{CoD}-\text{CoT})$ & +0.280 & +0.395 & \textbf{+0.455}\\
\midrule
Deepseek+CoT & 6.465 & 6.435 & 6.380\\
Deepseek+CoD & 6.580 & 6.560 & 6.530\\
$\Delta(\text{CoD}-\text{CoT})$ & +0.115 & +0.125 & \textbf{+0.150}\\
\bottomrule
\end{tabular}
\caption{The average scores of Baichuan2-7B and Deepseek-7B with different numbers of candidate articles (evaluated by GPT-4).}
\label{tab:num_of_evi}
\end{table}

\section{Rules to Identify the Article Reference}
\label{app:art_rule}

We examine whether the responses use an article as the basis by following rules:
\begin{itemize}
    \item If the article number appears in the response, we believe the LLM has used this article as a reference.
    \item We segment the responses into sentence~\cite{che-etal-2021-n} and calculate the longest common subsequence~(LCS) between each sentence and the article's content. If the length of the longest LCS exceeds one-third of the article, we believe the LLM has referenced this article.
    \item Otherwise, the article is considered not to serve as a reference.
\end{itemize}

\section{More Analysis}

\subsection{Lengths of Generated Responses}
\label{sec:output_lens}
In Table~\ref{tab:out_lens} we provide the average lengths of generated responses from each LLM in different settings. We find that the lengths of responses under different experimental settings can be similar. After checking the correlation coefficient between the difference in response lengths and the difference in GPT-4 scores, we believe that the GPT-4 evaluator does not show preference for longer responses.

\begin{table}[t]
\small
\centering
\setlength\tabcolsep{3pt}
\begin{tabular}{c|c|c}
\toprule
Model &	Avg. Lens. of base	& Avg. Lens. of CoD  \\
\midrule
Baichuan2-7B & 201.20 & 200.78\\
Deepseek-7B	& 214.29 & 219.09\\
Qwen-7B & 236.55 & 259.03\\
Xverse-7B & 236.61 & 244.60\\
\bottomrule
\end{tabular}
\caption{The average lengths of responses generated by different LLMs in baseline setting and CoD setting.}
\label{tab:out_lens}
\end{table}
\subsection{ Examples required Reasoning with Commonsense Knowledge}
\label{app:analysis_of_commonsense}
\begin{table}[t]
\centering
\begin{small}
\begin{tabular}{p{0.97\columnwidth}}
\toprule
\textbf{Question:} Do you still need to pay child support after having your guardianship revoked?\\
\textbf{Article:} \textit{Article 37} \red{Parents}, children, and spouses \red{who support the wards in the form of child support}, support for elderly parents, or spousal support in accordance with the law \red{shall continue to perform such obligations after they are disqualified by the people's courts as guardians.}\\
\textbf{Original analysis:} \textit{Article 37} explicitly stipulates [the content of Article 37]. This article pertains to guardianship and child support, but since \hl{the question does not mention revocation by the People's Court}, \hl{this article should not be used as a basis.} \\
\midrule
\textit{Low-quality modification:}\\
\textbf{Revised analysis:} \textit{Article 37} explicitly stipulates [the content of Article 37]. This article pertains to guardianship and child support. However, \hl{the user does not explicitly say who revokes her/his guardianship.} Thus, \hl{this article should not be used as a basis.} \\
\midrule
\textit{High-quality modification:}\\
\textbf{Revised analysis:} Article 37 stipulates that \violet{the revocation of guardianship does not affect existing obligations to pay child support.} Thus, \violet{this article should be used as a basis.} \\ 
\bottomrule
\end{tabular}
\caption{\label{tab:bad_revise_case}
Failed and successful cases for revising evidence analysis. \red{Red texts} are the key basis of the question. \violet{Violet texts} are correct analysis, while the \hl{texts with yellow background} are hallucinated parts. 
}
\end{small}
\end{table}

As shown in the \textit{original analysis} of Table~\ref{tab:bad_revise_case}, LLMs tend to focus on the literal differences between questions and law articles.  We humans have the background knowledge that only courts have the authority to revoke guardianship, thus often omit this information in our questions.  But, LLMs often struggle with such differences between the questions and rigorous law articles, thus may not yield correct analysis.  

\subsection{Significant Test}
We apologize that we have not provided significance test. We have to admit that the GPT-4 evaluation is indeed too expensive. A single evaluation of all results under all settings can cost \$150. It is difficult for us to repeat the evaluation multiple times, since it may cost thousands of dollars.

In this work, we use a low temperature when generating. And we set the temperature to 0.0 for GPT-4 evaluation.

We randomly sample 20 examples and employ Baichuan2-7B to repeat the experiments of the baseline setting for 5 times with different seeds. We then use GPT-4 to score the responses of every run. The standard error of the average scores is 0.0245, which is much smaller than the improvement brought by CoD.

We also repeat the experiments on Baichuan2-7B for 5 times under the setting of CoD, using the same subset. The standard error of average scores yielded by GPT-4 is 0.0510, which is also smaller than the improvement brought by CoD.

\section{Prompts of Chain-of-Discussion}
\label{app:prodis}
\subsection{Prompt of Question Analysis}
To obtain the question analysis, we employ the prompts as below ([\textit{\_\_text\_\_}] indicates the English translation of the prompts in Chinese):

\zhsmall{你是一个民法领域的专家，你需要从法律专业的角度分析一名咨询者提出的问题涉及哪些关键点。在分析问题之后，你还要分析检索器提供的参考法条是否能作为分析该问题的依据。请你紧紧围绕咨询者的问题进行分析，不要过度设想潜在的、与问题不相关的场景。}

{\small [ \textit{You are an expert in the field of civil law. You need to analyze the key points of a question posed by a consultant from a legal professional's perspective. After analyzing the question, you should also assess whether the reference articles provided by the retriever can serve as a basis for analyzing the issue. Please focus closely on the consultant's question and avoid overly imagining potential scenarios unrelated to the issue.}]}

\zhsmall{咨询者的问题是“\{\{question\}\}”}

{\small[ \textit{The consultant's question is:} “\{\{question\}\}”]}

\zhsmall{下面是检索器提供的参考法条：
\{\{articles\}\}
}

{\small [ \textit{Below are the reference articles provided by the retriever:} \{\{articles\}\} ]}

\zhsmall{接下来，请你分析咨询者的问题“\{\{question\}\}”请你站在咨询者的角度，首先分析咨询者所处的环境及其面对的客观事实，再围绕着咨询者的问题，分析他可能需要了解哪些方面的法律规定。你需要先对问题进行分析，再分析各个参考法条是否有该问题有关。请你遵循格式，以“问题分析：”、“法条分析：”分别作为两段分析的开头。}

{ \small [ \textit{Next, please analyze the consultant's question} "\{\{question\}\}" \textit{Please put yourself in the consultant's shoes, first analyzing the environment they are in and the objective facts they are facing. Then, focusing on the consultant's question, analyze which aspects of the legal articles they might need to understand. You need to first analyze the question, then assess whether each reference article is relevant to the issue. Please follow the format, starting each section of the analysis with "Question Analysis:" and "Article Analysis:" respectively.
}]
}

Since we find the LLMs tend to provide the article analysis accompanied with question analysis. We ask it to generate these two analyses separately. We stop the generation after "\zhsmall{法条分析：}" ([Article Analysis]) has been output.

\subsection{Prompt of Summary}

To summarize the question analyses from multiple LLMs, we use following prompt:

\zhsmall{你是一名法律专家，你需要评价数位律师针对某个法律咨询问题的分析，并给出最终的总结性分析。这些律师的分析既包含正确的内容，也包含错误的内容。你可以参考法条中的内容，谨慎判断各位律师对问题本身的分析是否正确，以及对潜在情况的讨论是否合理。请注意，多数律师都提及的内容更有可能是正确的，你在最终总结时可以参考这部分内容。而如果某个内容仅被个别律师提及，那么该内容有可能是错误的或者与问题不相关的，你需要谨慎判断它是否合理。}

{\small[ \textit{You are a legal expert. You need to evaluate several lawyers' analyses of a specific legal consultation question and provide a final summary analysis. These lawyers' analyses contain both correct and incorrect elements. You can refer to the content of the legal articles, carefully judging whether each lawyer's analysis of the issue itself is correct and whether their discussion of potential scenarios is reasonable. Note that content mentioned by most lawyers is more likely to be correct, and you can reference this part in your final summary. If a particular point is mentioned by only a few lawyers, it may be incorrect or irrelevant to the issue, so you need to carefully judge its reasonableness.
}] }

\zhsmall{你需要先对律师们的分析进行点评，你需要评价这些分析中的每一个要点是否正确、是否与问题相关。对于存在逻辑错误或者与问题相关度较低的要点，你要明确指出并给予批评。之后，请你基于你的点评，给出一段语气、句式都与各位律师对该问题的分析相似的总结性分析。}

{\small [ \textit{You first need to critique the lawyers' analyses, evaluating whether each point in these analyses is correct and relevant to the issue. For points with logical errors or low relevance to the issue, you should clearly point them out and provide criticism. Afterward, based on your critiques, provide a summary analysis that matches the tone and style of the lawyers' analyses of the issue.
}]
}

\subsection{Prompt of Evidence Analysis}

To obtain analyses for articles, we use following prompts:

\zhsmall{你是一名法律专家，你需要判断某个法条是否能作为依据，用于解答给定的法律咨询问题。针对咨询者提出的问题，检索器提供了五个参考法条。但是这些法条可能对于解答问题有帮助，也可能没有帮助。你需要逐个对法条进行分析和判断，在针对某个法条进行判断时，请你不要对其他法条进行判断。}

{\small [ \textit{You are a legal expert tasked with determining whether a specific legal article can serve as a basis for answering a given legal consultation question. The retriever has provided five reference articles for the question posed by the consultant. However, these articles may or may not be helpful in answering the question. You need to analyze and evaluate each article individually, refraining from making judgments about other articles when assessing a particular one.
}]}

\zhsmall{咨询者的问题是“\{\{question\}\}”}

{\small [ \textit{The consultant's question is:} \{\{question\}\} ]}

\zhsmall{检索器提供的参考法条：
\{\{articles\}\} }

{\small [ \textit{The reference articles provided by the retriever are:} \{\{articles\}\} ]}

\zhsmall{你需要先对该问题的关键点进行分析，然后再逐个分析每个法条是否对于解答该问题有帮助。分析法条的过程中，请你先思考法条规定了何种权利和义务，或者对何种行为实施了禁止令。如果法条中规定或禁止的内容与问题中的关键点有一定的相关性，那么该法条有可能对于解答问题有所帮助；否则，该法条大概率对解答问题没有帮助。}

{\small [ \textit{You need to first analyze the key points of the question and then proceed to analyze each article to determine its relevance to answering the question. During the analysis of the articles, consider what rights and obligations the article specifies, or what behaviors it prohibits. If the articles are somewhat relevant to the key points of the question, then the article may likely be helpful in answering the question; otherwise, the article is probably not helpful in addressing the question.}]
}

\subsection{Prompt of Critique}

When criticize other LLM's article analysis, we use following prompts:

\zhsmall{你是一名法律专家，你需要点评一名律师对于某个法条是否有助于解答某个法律咨询问题的分析是否误解了法条的内容。我会明确告知你问题和法条的具体内容。}

{\small[ \textit{You are a legal expert. You need to critique a lawyer's analysis of whether a specific legal article is helpful in answering a particular legal consultation question and determine if their analysis misconstrues the content of the legal article. I will provide you with the specific details of the question and the legal article.
}] }

\zhsmall{问题：\{\{question\}\} }

{\small [ \textit{Question: }\{\{question\}\} ]}

\zhsmall{法条：\{\{article\}\}}

{\small [ \textit{Articles: }\{\{article\}\} ] }

\zhsmall{律师对于法条的分析：\{\{art\_ana\}\}}

{\small [ \textit{Lawyer's analysis for the articles:} \{\{art\_ana\}\} ] }

\zhsmall{接下来，请先用简洁的语言点评律师对于\{\{cur\_art\_id\}\}的分析。之后，请你判断他的分析是否误解了法条的内容。}

{\small [ \textit{Next, please provide a concise critique of the lawyer's analysis of} Article No. \{\{cur\_art\_id\}\}. \textit{Then, determine whether their analysis misconstrues the content of the legal article.
}]
}

\subsection{Prompt of Revising}

When revising, we employ following prompts:

\zhsmall{你是一名律师，你对于某个法条是否有助于解答某个法律咨询问题进行了点评。一些法学专家认为你的点评中存在对法条内容的理解、法条与问题之间的关联性等角度存在错误。你需要参考你对问题的分析，修改你对法条的分析。}

{\small [ \textit{You are a lawyer who has provided an assessment of whether a certain legal article is helpful in addressing a specific legal consultation question. Some legal experts believe there are errors in your assessment regarding understanding the content of the legal article and its relevance to the question. You need to revise your analysis of the legal article based on your analysis of the issue.
}] }

\zhsmall{问题：\{\{question\}\}}

{\small [ \textit{Question:} \{\{question\}\} ]}

\zhsmall{法条：\{\{article\}\}}

{\small [ \textit{Articless}: \{\{article\}\} ]}

\zhsmall{律师对于问题的分析：\{\{que\_ana\}\}}

{\small[ \textit{Lawyer's Analysis of the Question: }\{\{que\_ana\}\} ] }

\zhsmall{师对于法条的分析：\{\{art\_ana\}\} }

{\small [ \textit{Lawyer's Analysis of the Legal Article: }\{\{art\_ana\}\} ]}

\zhsmall{专家点评：\{\{critiques\}\}}

{\small [ \textit{Expert Critiques: }\{\{critiques\}\} ]}

\zhsmall{接下来，请你重写一份更为正确的法条分析。在重写后的法条分析的结尾，请你按照你的分析，评估一下该法条是否可能有助于解答问题。}

{\small[ \textit{Next, please rewrite a more accurate analysis of the legal article. At the end of the rewritten analysis of the legal article, evaluate whether the legal article may indeed be helpful in addressing the question based on your analysis.
}]
}

\subsection{Prompt of Response}

Finally, to response to the user's question, we use following prompts:

\zhsmall{
你是一个法律专家，你需要从法律专业的角度回答咨询者提出的问题。你需要以具体的法条为依据回答问题，并告诉咨询者法律赋予他哪些权利和义务，或者禁止他实施哪些举措。在回答问题之前，你可以参考检索器提供的一些参考法条。但请注意，检索器提供的法条并不一定都有助于回答咨询者提出的问题，它也可能与提问者的问题无关。因此，你需要对问题涉及的事实背景进行分析，再判断各个法条是否能够作为回答问题的依据。请你不要将检索器提供的全部参考法条都当作依据，也不要引用参考法条之外的其他法条作为依据。在回答的过程中，请你紧紧围绕提问者的问题进行讨论，不要过度设想潜在的、与问题不相关的情形。}

\zhsmall{[ \textit{You are a legal expert, and you need to answer the question posed by the consultant from a legal perspective. You are required to provide specific legal articles as the basis for your answer, informing the consultant of their rights, obligations conferred by the law, or actions prohibited by it. Before answering the question, you can refer to some reference articles provided by the retriever. However, please note that the articles provided by the retriever may not necessarily be helpful in answering the consultant's question; they may also be irrelevant to the question. Therefore, you need to analyze the factual background of the issue involved, then determine whether each article can serve as a basis for answering the question. Please do not consider all the reference articles provided by the retriever as the basis, nor cite any articles outside the reference ones as evidence. During your response, focus closely on the consultant's question, avoiding overly imagining potential scenarios unrelated to the issue.}]}

\zhsmall{
咨询者的问题是“\{\{question\}\}”}

\zhsmall{[ \textit{The consultant's question is: }\{\{question\}\} ]}

\zhsmall{
下面是检索器提供的参考法条：
\{\{articles\}\} }

\zhsmall{[ \textit{Below are the reference articles provided by the retriever: }\{\{articles\}\} ]}

\zhsmall{
接下来，请你回答咨询者提出的问题“\{\{question\}\}”
你需要先对该问题的关键点进行分析，然后判断各个参考法条是否有助于解答该问题。最后请你使用与该问题有关的部分法条作为依据，给出详细的回答。回答过程中禁止使用参考法条之外的内容。}

\zhsmall{[ \textit{Next, please answer the question posed by the consultant} "[[question]]" \textit{You need to analyze the key points of the question first, then determine whether each reference article is helpful in answering it. Finally, please provide a detailed answer using relevant portions of the articles as the basis. Use of content outside the reference articles is prohibited during the response.
}]}

\zhsmall{
问题分析：\{\{que\_ana\}\} }

\zhsmall{[ \textit{Question Analysis}: \{\{que\_ana\}\} ]}

\zhsmall{
法条分析：\{\{art\_ana\}\} }

\zhsmall{[ \textit{Article Analysis}: \{\{art\_ana\}\} ]}

\zhsmall{
回答：}

\zhsmall{[ \textit{Answer:}]}

\section{Scoring Prompt of GPT-4}
\label{app:gpt4_scoring}

Following CritiqueLLM~\cite{ke2023critiquellm}, we employ a reference-based prompt to instruct GPT-4 to assess the overall quality of the responses generated by open-source LLMs. We use the human-written response and the \textit{necessary} and \textit{option} articles as reference for evaluation. The prompt is shown as below: 

{\small
[Instruction]

Please act as an impartial judge and evaluate the quality of the response provided by an AI assistant to the user question displayed below. Your evaluation should consider factors such as the logicality, helpfulness, relevance, accuracy, depth, and whether using irrelevant articles beyond the reference articles as a basis. Begin your evaluation by providing a short explanation. You will be given several reference articles, a high-quality reference answer and the assistant’s answer. Be as objective as possible. You should first provide your explanation IN CHINESE, then you must rate the response on a scale of 1 to 10 by STRICTLY following the below MAPPING for the relation between the scores and response quality:

1. The score 1\textasciitilde 2 stands for very chaotic or absence of answer, and the AI assistant completely failed to answer the user’s question. Serious logical and factual errors might also be included in this term. The gap between the AI assistant’s answer and the high-quality reference answer is huge and insuperable.

2. The score 3\textasciitilde 4 indicates fragment-like responses from AI assistant’s answer. It did not provide answers in proper grammar, fluency, or accuracy. Citing irrelevant articles and resulting in a redundant output also falls under this scenario. There are obvious gaps between the high-quality reference answer and the AI assistant’s response.

3. The score 5\textasciitilde 6 indicates for existence of minute disadvantage from the AI assistant’s answer compared to the high-quality reference answer. Yet the AI assistant did provide an average answer. The AI assistant either did not fully address the question, or was somewhat short of logicality, helpfulness, relevance, depth, or detailedness. The disadvantages from the AI assistant’s answer overwhelm its advantages.

4. The score 7\textasciitilde 8 indicates the AI assistant provided a good answer as well as the high-quality reference answer, addressing the question, with good helpfulness, relevance, accuracy, depth, creativity, and enough details. The response of AI assistant does not include any irrelevant articles beyond the reference articles. The AI assistant might have flaws compared to the reference answer, but that does not overwhelm the above advantages.

5. The score 9\textasciitilde 10 indicates the AI assistant responded better than the provided reference answer in most aspects, fully achieved the instruction goal, provided more detailed analysis, and have unique advantages to the reference answer.

You should give scores around 7 if you do not find obvious advantages or disadvantages. You should seriously consider the above guide before give lowest and highest scores such as 1 or 10, and avoid such situation if you do not have sound explanations. Avoid any positional biases and ensure that the order in which the responses were presented does not influence your decision. Do not allow the length of the responses to influence your evaluation. Do not favor certain names of the assistants. AND again, VERY IMPORTANTLY, after you provide your explanation IN CHINESE, you must rate the response strictly following this format: “Rating: [[Number]]”, for example: Rating: [[5]].

[User’s Question]

\{\{QUESTION\}\}

[The Start of Reference Articles]

\{\{ARTICLES\}\}

[The End of Reference Articles]

[The Start of Reference Answer]

\{\{GOLDEN RESPONSE\}\}

[The End of Reference Answer]

[The Start of Assistant’s Answer]

\{\{LLM'S RESPONSE\}\}

[The End of Assistant’s Answer]
}

\end{document}